%% file: main.tex
\documentclass[10pt,twocolumn,letterpaper]{article}

\usepackage[pagenumbers]{cvpr} 
\usepackage{times}
\usepackage{caption}
\usepackage{epsfig}
\usepackage{graphicx}
\usepackage{amsmath}
\usepackage{amssymb}
\usepackage{bm}
\usepackage{xcolor}
\usepackage{array}
\usepackage{booktabs}
\usepackage{multirow}
\usepackage[accsupp]{axessibility}
\usepackage{lineno}

\definecolor{mygreen}{HTML}{008000}
\definecolor{myred}{HTML}{D10000}


\definecolor{cvprblue}{rgb}{0.21,0.49,0.74}
\usepackage[pagebackref,breaklinks,colorlinks,citecolor=cvprblue]{hyperref}

\def\paperID{10373} 
\def\confName{CVPR}
\def\confYear{2025}

\title{FreeUV: Ground-Truth-Free Realistic Facial UV Texture Recovery \\via Cross-Assembly Inference Strategy}

\author{Xingchao Yang$^{1,2}$, Takafumi Taketomi$^{1}$, Yuki Endo$^{2}$, Yoshihiro Kanamori$^{2}$\\
$^1$CyberAgent \quad$^2$University of Tsukuba\\
{\tt\small \{you\_koutyo, taketomi\_takafumi\}@cyberagent.co.jp, \{endo, kanamori\}@cs.tsukuba.ac.jp}
}

\begin{document}
\twocolumn[{%
\maketitle
\begin{center}
    \centering
    \captionsetup{type=figure}
    \includegraphics[width=\textwidth]{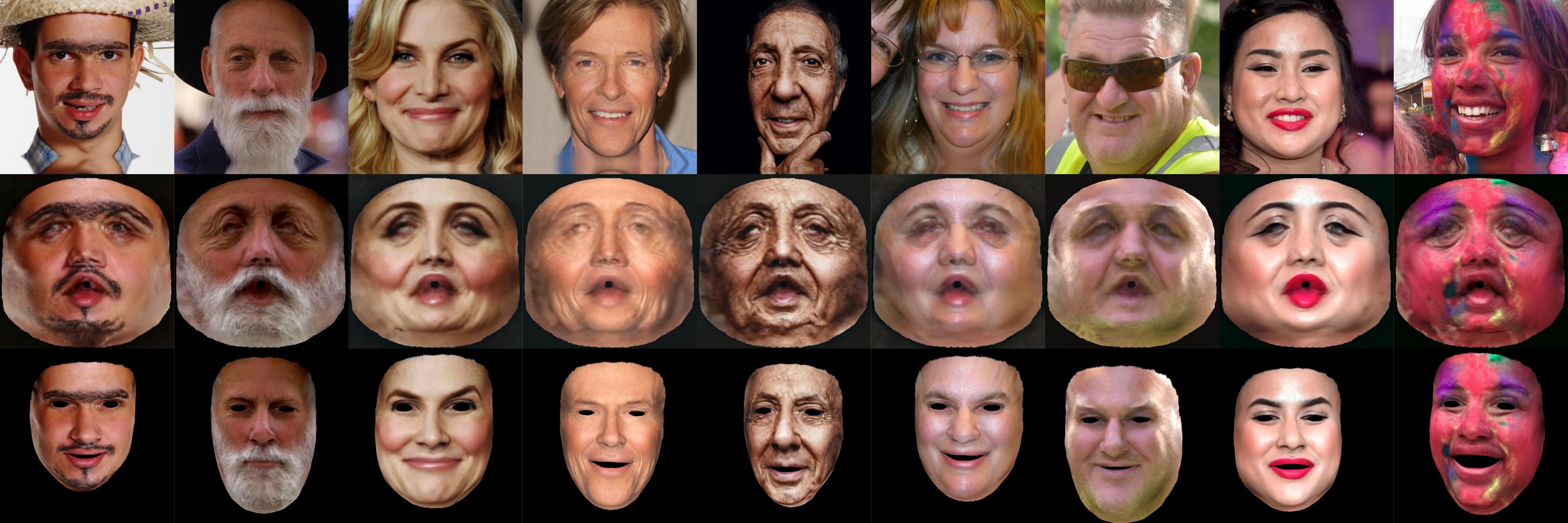}
    \captionof{figure}{\textbf{Examples of FreeUV results.} Top to bottom: input face images, recovered UV textures, and FLAME model-based rendering. FreeUV generates a complete UV texture from a single face image without requiring ground-truth UV supervision during training. The method captures intricate details, such as facial hair, wrinkles, occlusions, and makeup, while demonstrating robustness across diverse scenarios, achieving high fidelity and coherent texture recovery.}
    \label{fig:teaser}
\end{center}%
}]

\input{sec/0_submission}

\input{sec/X_suppl}

\clearpage
{
    \small
    \bibliographystyle{ieeenat_fullname}
    \bibliography{main.bib}
}

\end{document}

%% file: sec/0_submission.tex
\begin{abstract}
Recovering high-quality 3D facial textures from single-view 2D images is a challenging task, especially under the constraints of limited data and complex facial details such as wrinkles, makeup, and occlusions. In this paper, we introduce \textcolor{pink!60!black}{\textbf{FreeUV}}, a novel \textbf{ground-truth-free} UV texture recovery framework that eliminates the need for annotated or synthetic UV data. FreeUV leverages a pre-trained stable diffusion model alongside a Cross-Assembly inference strategy to fulfill this objective. In FreeUV, separate networks are trained independently to focus on realistic appearance and structural consistency, and these networks are combined during inference to generate coherent textures. Our approach accurately captures intricate facial features and demonstrates robust performance across diverse poses and occlusions. 
Extensive experiments validate FreeUV's effectiveness, with results surpassing state-of-the-art methods in both quantitative and qualitative metrics. Additionally, FreeUV enables new applications, including local editing, facial feature interpolation, and texture recovery from multi-view images. 
By reducing data requirements, FreeUV offers a scalable solution for generating high-fidelity 3D facial textures suitable for real-world scenarios.
\end{abstract}

\section{Introduction}
\label{sec:intro}

Reconstructing realistic 3D facial textures from single 2D images is a longstanding challenge in computer vision and graphics. A major breakthrough in this field came with the introduction of the 3D Morphable Model (3DMM) by Blanz and Vetter~\cite{Blanz3DMM99}, which enabled parametric modeling of 3D face shapes and allowed for effective 3D facial reconstruction from limited 2D input. Since then, 3DMMs have been significantly refined and expanded, resulting in a variety of models that improve reconstruction accuracy, robustness, and adaptability~\cite{bfm, LYHM, lsfm1, lsfm2, FLAME, CoMA, DiverseRaw, FML, AlbedoMM, FaceScape, CompleteMM, 3DMMfromSingle, FaceVerse, nonlinear3DMM, LearningPhyFace, really2022, i3DMM, headnerf, MoFaNeRF, NeuralHeadModel, TRUST, ASM, egger20203d, FaceWarehouse, IlluminationPrior, GlobalIllumiMM, AlbedoMM, MakeupExtract, makeupPriors2024, LAMM2024}. Despite these advances, achieving high-quality textures that faithfully capture complex facial details—such as wrinkles, pores, facial hair, and diverse makeup—continues to be a challenge, particularly when only single-view, in-the-wild images are available.

To achieve high-quality 3D facial texture generation under constrained resources and lower costs, recent approaches~\cite{saito2017photorealistic, HFFRI, GANFIT, uvgan, LearningPhyFace, OSTeC, AvatarMe, AvatarMe++, UnsupervisedHFFG22, tbgan2022, NormalizeAvatar, bareskinnet, FitMe, PracticalFaceReconsCGF, FFHQ-UV, relightify2023, makeupPriors2024, id2reflect2024, uvidm2024, mosar2024} have increasingly turned to generative models, such as GANs, to synthesize realistic textures. These generative models utilize non-linear structures to capture complex textures, achieving impressive levels of realism from a single face image. However, most 3D facial UV texture generation methods rely on supervised learning, which requires ground-truth UV datasets. 

Facial UV texture generation methods can be broadly divided into two main categories based on the type of training data used: methods that use captured real data and methods that rely on synthesized data. 
The first category relies on costly UV datasets captured with specialized equipment as ground truth but lacks generalization to in-the-wild conditions~\cite{saito2017photorealistic, HFFRI, LearningPhyFace, AvatarMe, AvatarMe++}. 
The second category involves generating synthetic UV datasets using models like StyleGAN~\cite{StyleGAN, StyleGAN2} to create training data for facial texture models~\cite{FFHQ-UV, mvinvert2022, uvidm2024}. This approach faces two main limitations: (1) it is constrained by the capabilities of StyleGAN, often resulting in domain limitations that make it difficult to handle diverse, unseen faces, such as those with makeup; and (2) StyleGAN-based UV texture generation typically involves multiple steps. First, GAN inversion~\cite{GANInversion} is used to create multi-view images by adjusting the facial pose. These images are then blended to produce a complete UV texture. However, this process often results in inconsistencies in identity, expression, lighting, and appearance, making it difficult to generate realistic and coherent textures across views.
In summary, both categories are fundamentally constrained by their dependence on large-scale, realistic UV textures—real or synthetic—as ground truth, which is essential for effective model training.

\begin{figure}[t]
    \centering
    \includegraphics[width=\linewidth]{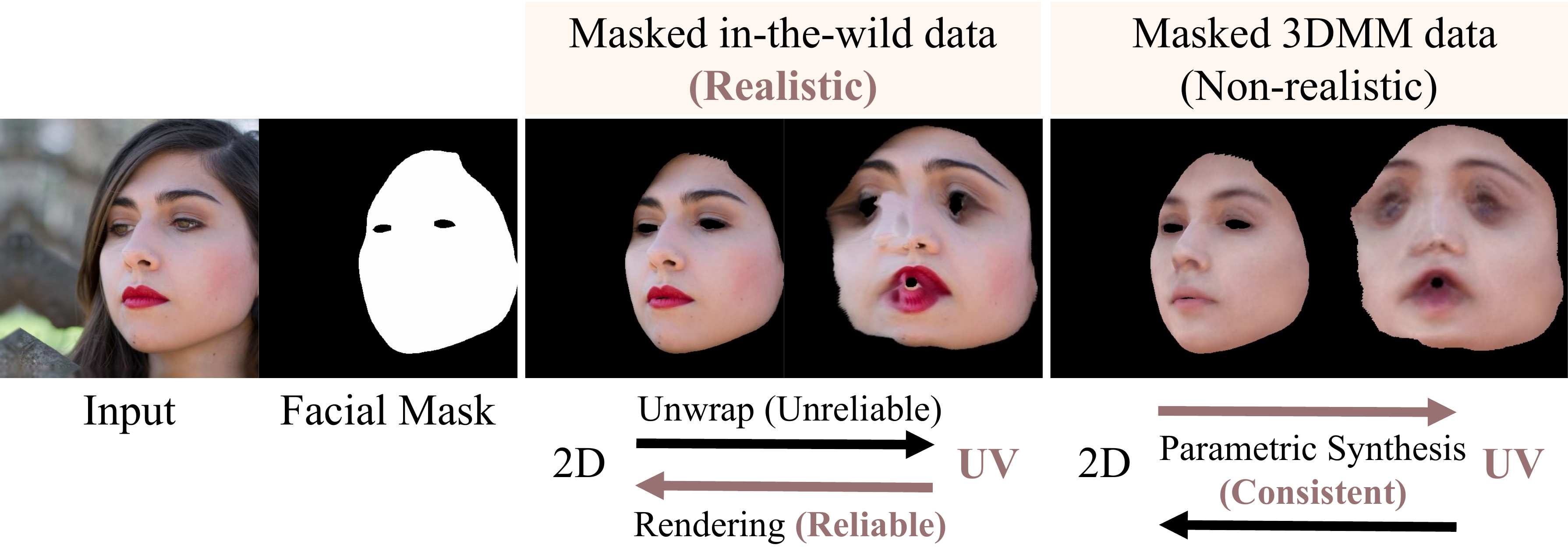}
    \caption{\textbf{Example of data and domain characteristics used in FreeUV.} Realistic textures are derived from in-the-wild data, while structurally consistent textures are generated from parametric 3DMM data.
}
\label{fig:domain_select}
\end{figure}

In this paper, we introduce FreeUV, a novel \textbf{ground-truth-free UV} texture generation framework that recovers realistic 3D facial UV textures from single 2D images (\cref{fig:teaser}). 
Unlike many existing methods, 
FreeUV does not require complete UV ground truth during training, regardless of whether real or synthetic.
Instead, it utilizes a pre-trained stable diffusion model~\cite{stablediffusion2022} and a unique Cross-Assembly inference strategy. We independently train appearance feature extraction and structural reconstruction modules, which are then integrated during inference. This setup enforces a disentanglement of appearance and structure, enabling the model to learn robust texture representations directly from partial or highly misaligned flaw UV texture inputs (see \cref{fig:domain_select}, third column), thereby eliminating the need for fully annotated UV datasets. By minimizing data dependency, FreeUV significantly reduces the cost and complexity of generating high-quality UV textures.

Our main idea, as illustrated in \cref{tab:domain_comparison} and \cref{fig:domain_select}, is to selectively leverage the strengths of both in-the-wild real images and related 3DMM data by combining their functionalities. Specifically, during the training stage, FreeUV uses two networks, each with a complementary role:
(1) The appearance feature extraction network prioritizes realism by focusing on the in-the-wild domain. Trained to map degraded UV textures to masked 2D face images, this network utilizes a CLIP-based~\cite{CLIP2021} feature extractor with channel attention~\cite{channelattn2018} to capture fine facial details while ensuring structural coherence.
(2) The structural reconstruction network emphasizes structural control independently of appearance. Leveraging ControlNet~\cite{controlnet23}, it operates in the 3DMM domain, mapping masked 3DMM face images to masked 3DMM UV textures to establish structural consistency.
In the inference phase, we integrate the UV-specific modules from both networks into a pre-trained stable diffusion model, generating realistic, coherent UV textures. FreeUV effectively captures intricate details such as facial hair, wrinkles, and makeup while maintaining robustness across various poses and occlusions. By eliminating the need for ground-truth UV data, FreeUV provides a scalable, data-efficient approach to high-quality 3D facial texture generation for real-world applications.

\begin{table}[t]
\caption{\textbf{Selective domain utilization in FreeUV’s texture recovery.} Here we illustrate the Cross-Assembly strategy in FreeUV, highlighting how realistic appearance from in-the-wild images and structural consistency from 3DMM are selectively combined. The final setup targets a \textbf{\textcolor{pink!60!black}{UV-to-UV}} mapping with a \textbf{\textcolor{pink!60!black}{Realistic and Consistent}} combination for optimal texture generation.}

\begin{center}
\resizebox{\linewidth}{!}{
\begin{tabular}{>{\centering\arraybackslash}p{1.6cm} l|cc|c}
\toprule
\textbf{Mapping} & \textbf{Domain} & \textbf{Appearance} & \textbf{Structure} & \textbf{Selected} \\
\midrule
\multirow{2}{*}{\textbf{\textcolor{pink!60!black}{UV-}}to-2D} 
& 3DMM & Non-realistic & Consistent & \textcolor{myred}{×} \\
\cmidrule{2-5}
& In-the-wild & \textbf{\textcolor{pink!60!black}{Realistic}} & Reliable & \textcolor{mygreen}{\checkmark} \\
\midrule
\multirow{2}{*}{2D-to\textbf{\textcolor{pink!60!black}{-UV}}} 
& 3DMM & Non-realistic & \textbf{\textcolor{pink!60!black}{Consistent}} & \textcolor{mygreen}{\checkmark} \\
\cmidrule{2-5}
& In-the-wild & Realistic & \textcolor{gray}{Unreliable} & \textcolor{myred}{×} \\
\bottomrule
\end{tabular}
}
\end{center}
\label{tab:domain_comparison}
\end{table}

In summary, our contributions are:
\begin{itemize}
    \setlength{\itemsep}{0.2em}
    \setlength{\parskip}{0.2em}
    \item \textbf{FreeUV}: A framework for generating high-quality facial UV textures without the need for costly annotated data or large synthetic datasets. By minimizing data requirements, FreeUV offers a practical and scalable solution for real-world applications;
    \item \textbf{Cross-Assembly Inference Strategy}: We utilize two complementary networks that focus separately on feature extraction (UV to 2D) and structural reconstruction (2D to UV) during training. At inference, these modules are combined (UV to UV) to generate high-fidelity textures that align with the 3DMM UV layout, effectively reducing UV unwrapping distortions, especially in large-angle faces and self-occlusions;
    \item \textbf{Flaw-Tolerant Facial Detail Extractor}: A channel attention-enhanced facial detail extractor. This module accurately captures essential facial features—including facial hair, wrinkles, and makeup—while preserving structural integrity, significantly enhancing the quality and robustness of UV texture generation across diverse conditions.
\end{itemize}

\section{Related Work}

\subsection{3DMM-based Face Reconstruction}
3D Morphable Models (3DMMs) have served as foundational tools in reconstructing 3D facial geometry and texture from 2D images. Introduced by Blanz and Vetter~\cite{Blanz3DMM99}, the 3DMM framework represents faces as linear combinations of shape and texture components derived from a set of 3D face scans. This framework has evolved significantly over time, incorporating both optimization-based and deep learning-based approaches to improve reconstruction accuracy and efficiency~\cite{reconstruction, PixelRec, Face2Face, RegressionDeep, mofa, CNNRecons, unsupervised3dmm, 250Hz, deng2020accurate, INORig, CEST, MICA, CNNInverse, TRUST, FOCUS, 3dffav32023, MonocularRTA, egger20203d, AccurateTransformer}. 
These methods typically rely on optimizing or regressing linear parameters within the 3DMM framework, which limits their expressive power. Some approaches specifically focus on enhancing geometry~\cite{DetailedRecons, DECA, EMOCA, FaceScape, hiface2023} to address this limitation. Iterative methods, such as HRN~\cite{hrn2023} and NextFace~\cite{PracticalFaceReconsCGF}, employ a coarse-to-fine refinement process to simultaneously improve texture and geometry, achieving more accurate reconstructions. While these geometry-focused approaches improve structural accuracy, they often fall short in capturing the detailed textures essential for realistic facial representation.

\subsection{High-Quality Facial Texture Recovery}
Beyond enhancements within the 3DMM framework, another research direction focuses on methods for directly generating high-quality facial textures aligned with the 3DMM UV layout. Some approaches leverage image translation method~\cite{image2image} trained on large-scale scanned datasets to infer high-resolution textures~\cite{saito2017photorealistic, HFFRI, AvatarMe, LearningPhyFace, AvatarMe++, Bao:2021:ToG}. However, these models are often resource-intensive, limiting their practicality.

Alternative methods employ decoder-based approaches~\cite{StyleGAN2, GANInversion} to produce detailed textures directly from 2D images~\cite{UnsupervisedHFFG22, THFFSOR, uvgan, GANFIT, OSTeC, dsd-gan, NormalizeAvatar, mvinvert2022, FFHQ-UV, makeupPriors2024, uvidm2024}, offering a practical solution for texture generation from in-the-wild image datasets. FFHQ-UV~\cite{FFHQ-UV} proposes a method for generating face-normalized UV maps but relies on a resource-intensive iterative refinement process. Makeup Prior Models~\cite{makeupPriors2024} similarly use iterative optimization to create facial makeup UV textures. UV-IDM~\cite{uvidm2024}, in contrast, generates multi-view images first, synthesizing paired 2D face images and UV maps to train a diffusion-based generative model.

While these methods improve facial UV texture quality, their dependency on StyleGAN-based frameworks restricts their generative scope and requires intricate, multi-step workflows, often leading to inconsistencies and difficulties with diverse, previously unseen faces. FreeUV, in contrast, circumvents these limitations by eliminating the need for StyleGAN-generated synthetic complete UV data and employing a diffusion-based framework. This approach enables FreeUV to produce high-quality, realistic UV textures, capturing fine details with enhanced robustness across varied facial attributes and conditions.

\begin{figure*}[!ht]
    \centering
    \includegraphics[width=\linewidth]{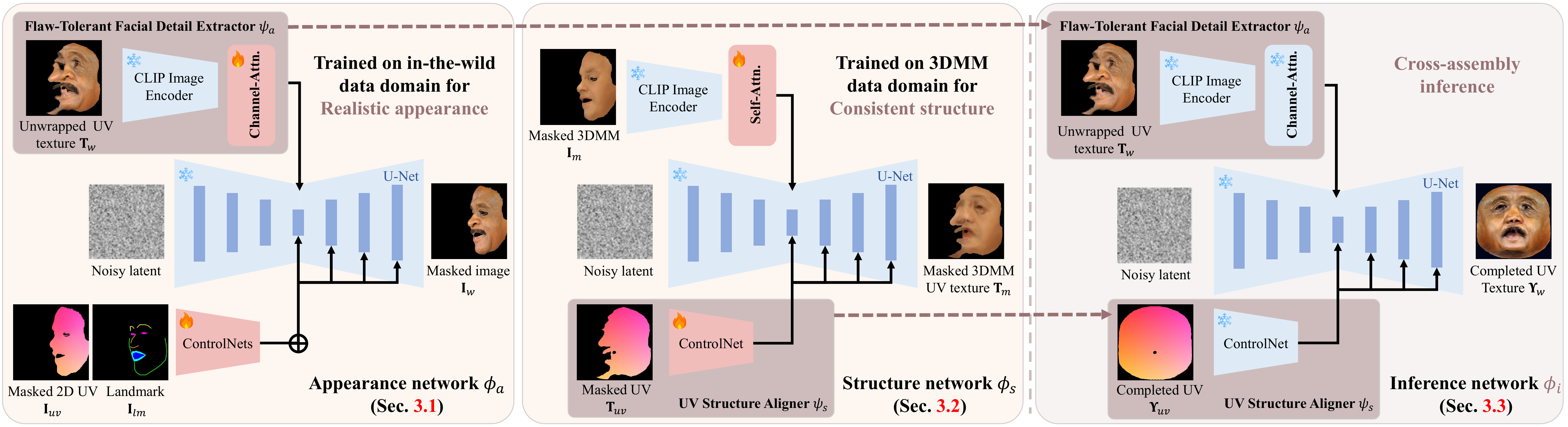}
    \caption{\textbf{Overview of FreeUV Framework.} FreeUV leverages two modules, the Flaw-Tolerant Detail Extractor ${\psi}_a$ (left) and the UV Structure Aligner ${\psi}_s$ (middle), to separately capture realistic appearance and structural consistency. Combined during the Cross-Assembly inference phase (right), these modules produce high-quality UV textures from single-view images, without requiring ground-truth UV data.}
\label{fig:method}
\end{figure*}

\subsection{Diffusion Models}
Diffusion models (DMs) are generative models that transform gaussian noise into target data distributions through an iterative denoising process, yielding realistic, high-quality images~\cite{diffusion, ddpm}. Due to their effectiveness, diffusion models have become the mainstream approach in image generation~\cite{ddim2021, textualinversion2023, dreambooth2023, plugandplay2023, ipadapter2023, T2I2024, sdxl2024}.
To enhance efficiency, the Latent Diffusion Model (LDM)~\cite{stablediffusion2022} encodes data into a compressed latent space, reducing computational costs while maintaining fidelity. LDMs, including Stable Diffusion (SD)~\cite{stablediffusion2022}, achieve the state-of-the-art results, particularly when trained on large-scale datasets. SD incorporates CLIP~\cite{CLIP2021}, a powerful text encoder, which further improves multimodal generation quality.
CLIP~\cite{CLIP2021} is a multi-modal model combining an image encoder and a text encoder to create aligned visual and textual representations. This alignment enables diffusion models, including SD, to leverage CLIP’s embeddings for image generation tasks~\cite{stablediffusion2022, ipadapter2023, T2I2024, SSREncoder2024, stablemakeup2024, stablehair2024}, expanding the flexibility and control in diffusion-based generation.
ControlNet~\cite{controlnet23} extends the functionality of diffusion models by introducing control signals from various modalities, such as depth or facial landmarks, enabling precise and controllable image generation. By utilizing a parallel U-Net structure with trainable layers that integrate spatial control conditions, ControlNet allows diffusion models to generate images guided by specific inputs without altering the core model.
Our method is based on Stable Diffusion (SD), leveraging an enhanced CLIP image encoder for feature extraction and ControlNet for structural control.

\section{Approach: FreeUV}
\label{sec:method_overview}
We start with an overview of FreeUV. Based on the insights from \cref{tab:domain_comparison} and \cref{fig:domain_select}, the overall framework of FreeUV is illustrated in \cref{fig:method}. During the training phase, FreeUV utilizes two separate networks—${\phi}_{a}$ and ${\phi}_{s}$—focusing on distinct aspects of appearance and structure. Specifically, ${\phi}_{a}$ is trained on in-the-wild data to enhance realistic appearance, while ${\phi}_{s}$ is trained on 3DMM data for structural consistency. In the inference phase, we employ a Cross-Assembly strategy to selectively integrate UV-specific modules from each network, capture both realistic appearance and consistent structure. The resulting integrated model, represented as ${\phi}_{i}$, to produce the final realistic completed UV texture.

A key contribution of FreeUV is that it does not require complete ground-truth UV textures during training. We use a pre-trained Stable Diffusion model~\cite{stablediffusion2022} as the backbone, with customized facial feature extractors based on the pre-trained CLIP encoder~\cite{CLIP2021} and apply ControlNet~\cite{controlnet23} for effective structure control. Our networks ${\phi}_{a}$ and ${\phi}_{s}$  leverage the image feature embedding approach proposed by \cite{ipadapter2023, SSREncoder2024, stablemakeup2024, stablehair2024}, integrating cross-attention layers within the U-Net architecture to capture detailed feature embeddings essential for realistic output. These networks are symmetrically structured: 
${\phi}_{a}$ performs a UV-to-2D mapping, while 
${\phi}_{s}$ performs a 2D-to-UV mapping. This symmetry in design helps avoid input-output similarity, reinforcing module disentanglement. In the Cross-Assembly inference phase, selected UV-specific modules are integrated into the pre-trained Stable Diffusion model~\cite{stablediffusion2022}, enabling conditional UV-to-UV texture generation.

To prepare the in-the-wild and 3DMM datasets, we employ a FLAME-based~\cite{FLAME} 3DMM fitting method~\cite{makeupPriors2024} focused on the face skin region. This method uses a re-trained version of the Deep3Dface~\cite{deng2020accurate} model for 3D face reconstruction. For an input image $\mathbf{I}$, we generate the paired data for training and inference.
Details on data preparation can be found in the supplementary materials.

In the following, we explain the two complementary networks: one focused on realistic appearance enhancement (\cref{sec:method_appearance}) and the other dedicated to maintaining structural consistency (\cref{sec:method_structure}). We then explain the Cross-Assembly inference strategy that integrates these modules for coherent texture generation (\cref{sec:method_inference}).

\subsection{Flaw-Tolerant Facial Detail Extractor} \label{sec:method_appearance}

In Appearance Network ${\phi}_{a}$, the Flaw-Tolerant Facial Detail Extractor ${\psi}_{a}$ comprises a CLIP image encoder with an additional channel attention layer. 
This module is designed to capture intricate facial details, while remaining robust against distortions and flaws typically present in unwrapped, in-the-wild UV data. To effectively leverage the CLIP image encoder, we adopted the CLIP feature extraction approach proposed in Stable-Makeup~\cite{stablemakeup2024}, gathering features from multiple layers of the CLIP visual backbone and concatenating them along the feature axis. Additionally, the channel attention mechanism~\cite{channelattn2018} selectively emphasizes relevant information while reducing the impact of less significant features. This combination enables ${\psi}_{a}$ to retain fine details and achieve high fidelity in the generated UV textures, even under challenging conditions.

Specifically, module ${\psi}_{a}$ extracts facial detail features from the unwrapped masked UV texture $\mathbf{T}_w$, selectively disregarding inaccurate features. 
Following structural cues from the masked rendered 2D UV position map $\mathbf{I}_{uv}$ and detected 2D landmarks $\mathbf{I}_{lm}$, the network generates the masked skin region image $\mathbf{I}_w$.
This entire process can be interpreted as a reliable UV-to-2D rendering.
We observe that 3DMM fitting does not achieve pixel-level alignment between $\mathbf{I}_{uv}$ and $\mathbf{I}_w$, and thus we incorporate 2D landmarks $\mathbf{I}_{lm}$ as an auxiliary alignment guide.

The loss function of network ${\phi}_{a}$ is similar to that of the Stable Diffusion model and can be formulated as follows:
\begin{equation}
\mathcal{L}_{a}(\theta) := \mathbb{E}_{\mathbf{x}_0, t, \boldsymbol{\epsilon}} \left[ \left\| \boldsymbol{\epsilon} - \epsilon_{\theta} (\mathbf{x}_t, t, \mathbf{c}_{T}^{w}, \mathbf{c}_{I}^{uv}, \mathbf{c}_{I}^{lm}) \right\|_2^2 \right],
\end{equation}
where
$\mathbf{x}_0$ is the original image latent, and $\mathbf{x}_t$ represents its noisy version at time $t$.
$t$ denotes the time step in the diffusion process.
$\boldsymbol{\epsilon}$ is the noise added to $\mathbf{x}_0$, which the denoising network $\epsilon_{\theta}$, 
with network parameters $\theta$,
aims to predict given $\mathbf{x}_t$.
$\mathbf{c}_{T}^{w}$, $\mathbf{c}_{I}^{uv}$, and $\mathbf{c}_{I}^{lm}$ are conditioning inputs embedded from $\mathbf{T}_w$, $\mathbf{I}_{uv}$, and $\mathbf{I}_{lm}$, respectively.

\subsection{UV Structure Aligner}
\label{sec:method_structure}

The UV Structure Aligner ${\psi}_{s}$ in structure network ${\phi}_{s}$ is designed to guide structural consistency that precisely aligns with the 3DMM UV layout. This network is trained using pixel-level aligned data derived from the same 3DMM parameters, including the masked 3DMM UV texture $\mathbf{T}_m$ and the masked UV position map $\mathbf{T}_{uv}$. In the data preparation stage, the masked rendered 3DMM image $\mathbf{I}_m$ is generated by rendering $\mathbf{T}_m$, ensuring data consistency. This entire process can be viewed as a consistent 2D-to-UV mapping.

For the feature extractor in structure network ${\phi}_{s}$, we adopt a CLIP based spatial-aware self-attention mechanism proposed by~\cite{stablemakeup2024}. 
In our experiments, we chose this distinct attention design for each mapping, based on the hypothesis that, due to the resolution difference, the UV-to-2D mapping acts as a “downsampling process,” where pixels are selectively sampled from the UV texture and mapped to the 2D image. Conversely, the 2D-to-UV mapping functions as an “upsampling process,” interpolating across missing details during unwrapping from the 2D image. 
Channel attention selectively identifies necessary features, while self attention captures relationships among features to enable accurate interpolation. Reversing these roles may introduce artifacts, as we will demonstrate in the experiments.

The loss function of network ${\phi}_{s}$ can be formulated as follows:
\begin{equation}
\mathcal{L}_{s}(\theta) := \mathbb{E}_{\mathbf{x}_0, t, \boldsymbol{\epsilon}} \left[ \left\| \boldsymbol{\epsilon} - \epsilon_{\theta} (\mathbf{x}_t, t, \mathbf{c}_{I}^{m}, \mathbf{c}_{T}^{uv}) \right\|_2^2 \right],
\end{equation}
where
$\mathbf{c}_{I}^{m}$, and $\mathbf{c}_{T}^{uv}$ are conditioning inputs embedded from $\mathbf{I}_m$, and $\mathbf{T}_{uv}$, respectively.

\subsection{Cross-Assemble Inference Strategy}
\label{sec:method_inference}

After separately training networks ${\phi}_{a}$ and ${\phi}_{s}$, we employ a Cross-Assembly strategy, combining ${\psi}_{a}$ and ${\psi}_{s}$ for inference. 
Specifically, ${\psi}_{a}$ is responsible for extracting realistic, detailed facial features from $\mathbf{T}_w$, while ${\psi}_{s}$ provides structural guidance based on the UV position map $\mathbf{\Upsilon}_{uv}$. Notably, we utilize the complete UV position map to ensure the recovery of the full UV texture $\mathbf{\Upsilon}_w$.

We observed that adjusting the classifier-free guidance scale during generation affects the color tone of the UV texture. This variation makes it challenging to align the color tone with the original image $\mathbf{I}_w$.
To address this, we apply a straightforward color transfer technique~\cite{colortransfer2001} as a post-processing step. This method adjusts the mean and standard deviation in the Lab color space of the generated UV texture, $\mathbf{\Upsilon}_w$, to match those of the image $\mathbf{I}_w$.

\section{Experiments}
\subsection{Implementation Details}
For the training dataset, we used a face segmentation method~\cite{BiSeNet2018} on the FFHQ dataset~\cite{FFHQ} to isolate face regions. After manual filtering to remove images with segmentation errors, hair, or occlusions, we selected 33,419 images for training.
For the Stable Diffusion backbone, we employed the pre-trained V1.5 model. The unwrapped texture used for training was resized to $512 \times 512$ pixels, and the generated UV textures also have the same resolution. The training was conducted on a single A100 GPU over 80,000 iterations, with a batch size of 4 and a learning rate of $3 \times 10^{-5}$. For inference, we utilized the DDIM sampler~\cite{ddim2021}, running 30 steps with a guidance scale of 1.4, which required 4.75 seconds per inference.

\subsection{Experimental Setup}
We utilized the high-resolution face databases FFHQ~\cite{FFHQ} and CelebAMask-HQ~\cite{CelebAHQMask2020}, selecting 10,000 images from each for both qualitative and quantitative evaluation. To further assess model performance under diverse conditions, we included 2,000 images randomly selected from the Large Pose Face Dataset (LPFF)~\cite{LPFF2023}. Through extensive experiments, we demonstrate FreeUV’s effectiveness and resilience in challenging scenarios. Specifically, our approach is compared against the state-of-the-art methods, including Deep3Dface~\cite{deng2020accurate}, FFHQ-UV~\cite{FFHQ-UV}, HRN~\cite{hrn2023}, UV-IDM~\cite{uvidm2024}, and Makeup Prior Models~\cite{makeupPriors2024}, where FreeUV consistently achieves superior performance across multiple metrics. Furthermore, FreeUV’s distinct design allows for flexible applications, supporting extended use cases in areas like customized local editing, facial feature interpolation, and multi-view texture recovery. Ablation studies were also conducted, providing a detailed qualitative and quantitative assessment of each module’s contribution to the overall effectiveness of FreeUV.

\subsection{Experiment Results}

\begin{figure}[t]
    \centering
    \includegraphics[width=\linewidth]{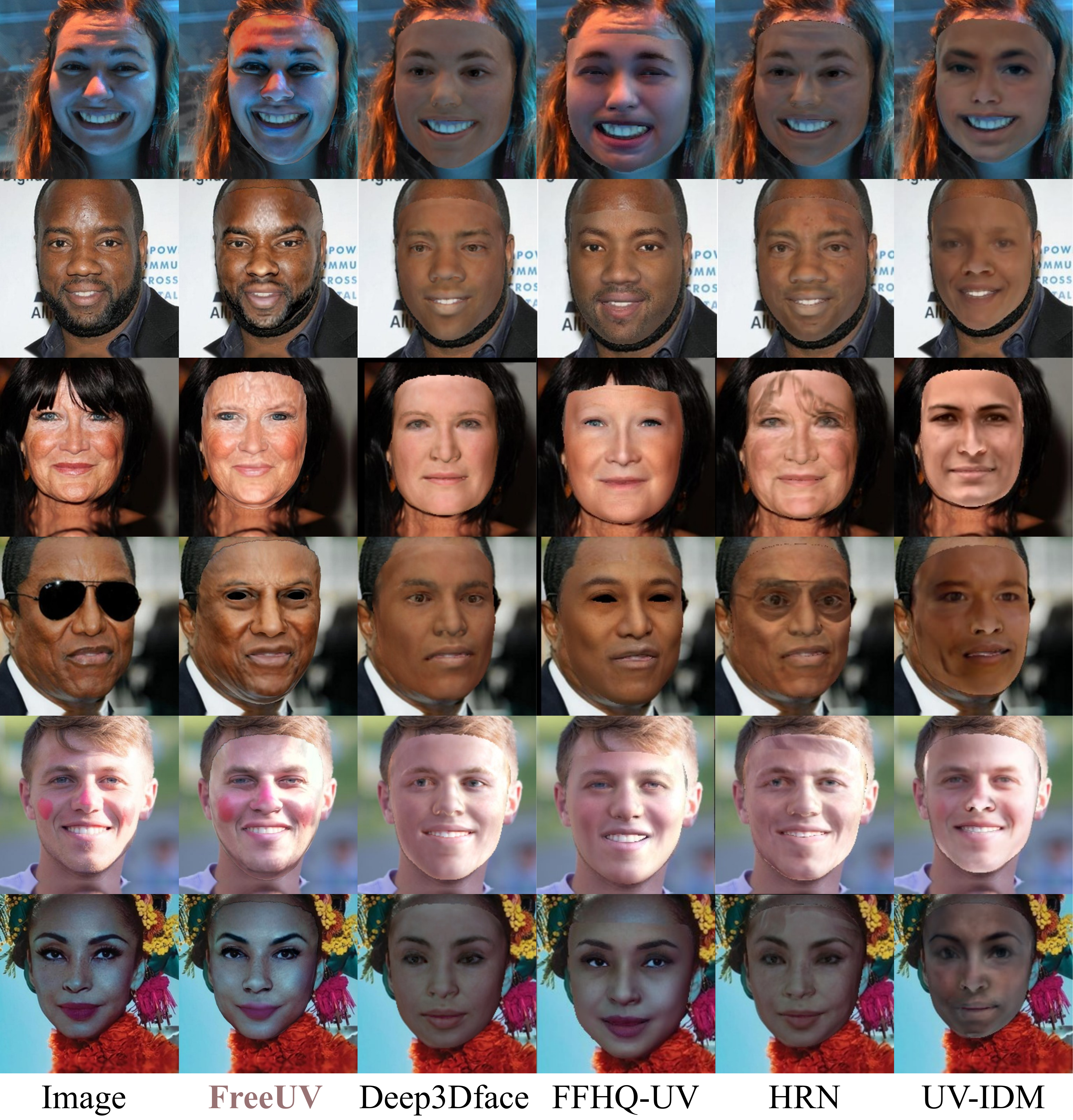}
   \caption{\textbf{Comparison of 3D face reconstruction results.} Our method achieves the closest match to the original input by rendering and overlaying the recovered UV texture. Even under challenging conditions, such as extreme lighting, facial hair, and occlusions, our approach preserves fine details and color consistency.
   }
\label{fig:result_recons}
\end{figure}

\begin{figure}[t]
    \centering
    \includegraphics[width=\linewidth]{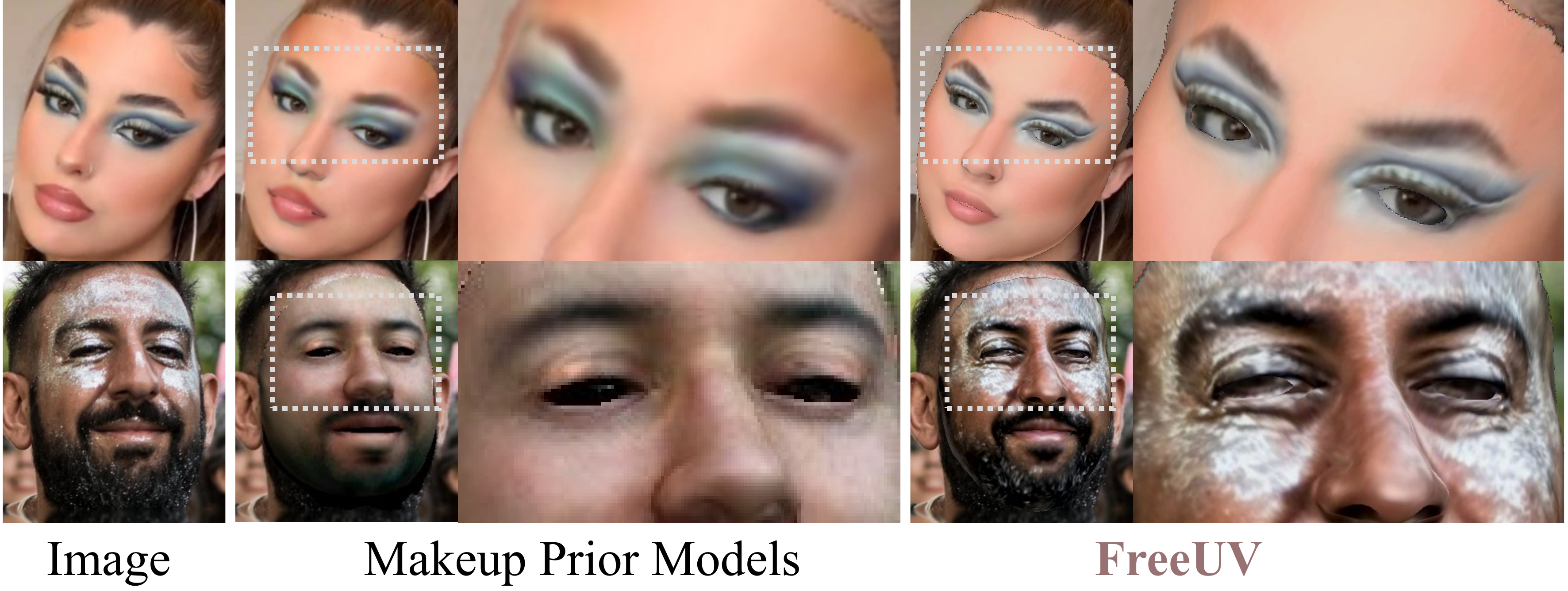}
    \caption{\textbf{Comparison with makeup-focused reconstruction method.} Our approach captures finer details, accurately preserving makeup features with greater clarity and consistency.
}
\label{fig:result_recons_makeup}
\end{figure}

\vskip0.5\baselineskip
\noindent \textbf{3D face reconstruction.}
To evaluate the quality of our recovered facial UV textures, we rendered and overlaid them onto original images for qualitative comparison in 3D face reconstruction. A quantitative analysis was not performed, as the comparison methods depend on 3DMM models~\cite{bfm, FLAME, hiface2023} and shape reconstruction techniques. 
As illustrated in Figs.~\ref{fig:teaser}, \ref{fig:result_recons}, and \ref{fig:result_recons_makeup}, our approach faithfully captures color variations and intricate facial details under challenging conditions, such as extreme lighting, specular highlights, facial hair, wrinkles, and makeup. 
Our method also demonstrates strong robustness against occlusions, including hair and glasses. In comparison to makeup-focused 3DMM techniques~\cite{makeupPriors2024}, our approach preserves fine makeup details—such as eyeliner—with exceptional fidelity.

\begin{figure}[t]
    \centering
    \includegraphics[width=\linewidth]{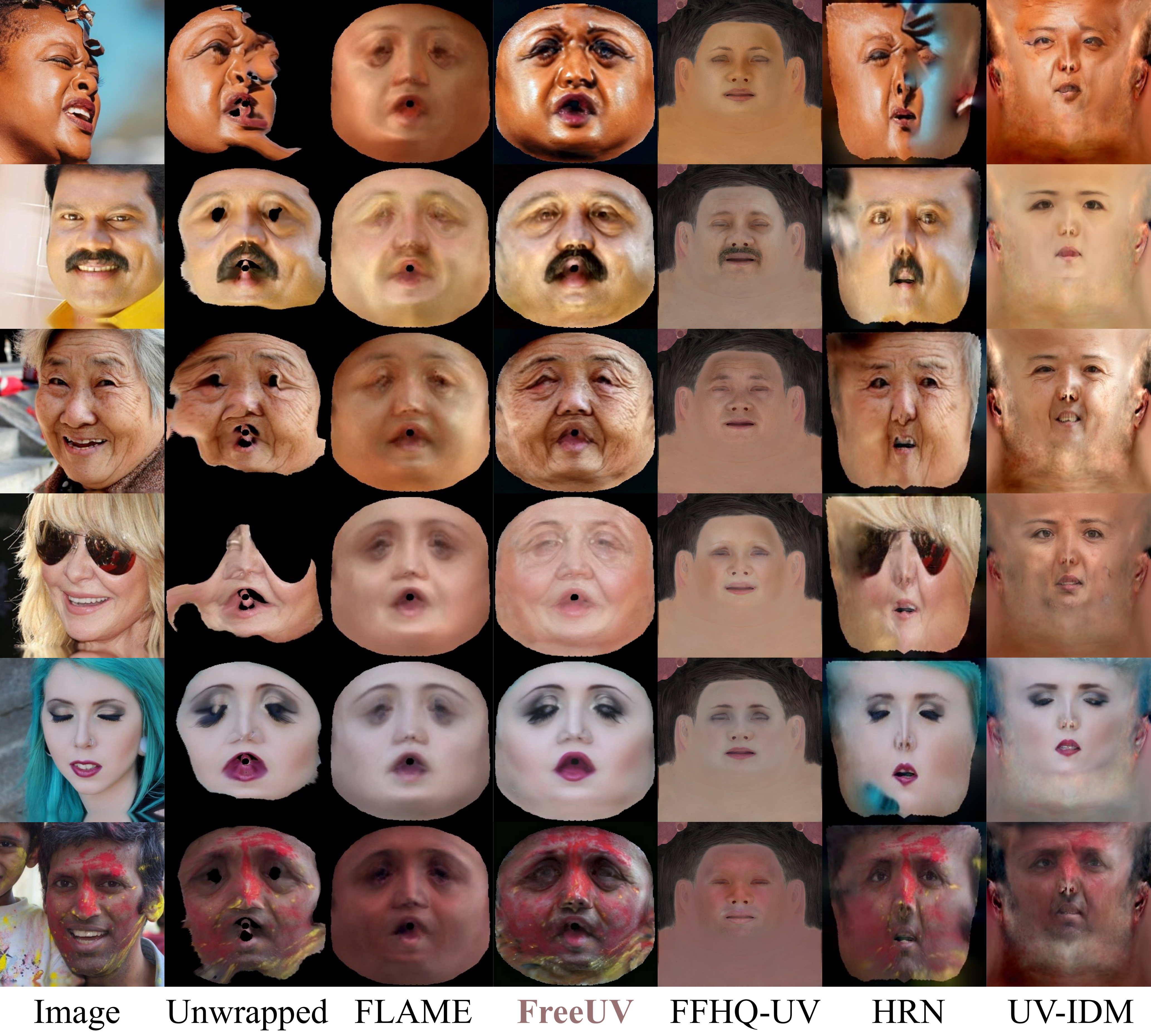}
    \caption{\textbf{Comparison of facial UV texture recovery.} Our method robustly produces realistic textures despite challenging inputs. Even with significant distortions, occlusions, and missing regions in the input data, the recovered UV textures retain fine details, smooth transitions, and consistent color tones.
}
\label{fig:result_uv}
\end{figure}

\vskip0.5\baselineskip
\noindent \textbf{UV texture recovery.}
As shown in \cref{fig:result_uv}, our method outperforms others by capturing finer details and significantly reducing artifacts. In the second column, despite inputs with substantial distortion, inaccuracies, and extensive missing regions, our approach reliably reconstructs realistic facial UV textures. We attribute this robustness to our Flaw-Tolerant Facial Detail Extractor, which selectively prioritizes accurate features, ensuring high-quality and detail-rich recovery.

\begin{table*}
\caption{\textbf{Comparative analysis of original 2D face images and the
recovered UV textures on FFHQ~\cite{FFHQ}, CelebAMask-HQ~\cite{CelebAHQMask2020}, and LPFF~\cite{LPFF2023} datasets.} Values in \textbf{bold} indicate the top-performing results, while those \underline{underlined} mark the second-best.}
\begin{center}
\resizebox{0.8\textwidth}{!}{
{\small
\begin{tabular}{lcccccccccc}
\toprule
\multicolumn{1}{l}{ } & \multicolumn{3}{c}{\textbf{FFHQ}} & \multicolumn{3}{c}{\textbf{CelebAMask-HQ}} & \multicolumn{3}{c}{\textbf{LPFF}} \\
\cmidrule(lr){2-4} \cmidrule(lr){5-7} \cmidrule(lr){8-10}
Method & CLIP-I$\uparrow$ & DINO-I$\uparrow$ & FID$\downarrow$ &  CLIP-I$\uparrow$ & DINO-I$\uparrow$ & FID$\downarrow$ & CLIP-I$\uparrow$ & DINO-I$\uparrow$ & FID$\downarrow$ \\ 
\midrule
HRN~\cite{hrn2023} & \underline{0.8327} & \underline{0.7389} & 166.19 & \underline{0.8259} & 0.7382 & 189.74 & 0.7368 & 0.5951 & \textbf{142.82} \\
UV-IDM~\cite{uvidm2024} & 0.7986 & 0.5836 & 228.74 & 0.7458 & 0.5690 & 258.34 & 0.7440 & 0.5345 & 239.10 \\
FLAME-based~\cite{makeupPriors2024} & 0.8218 & 0.7269 & \underline{158.06} & 0.8016 & \underline{0.7640} & \underline{164.98} & \underline{0.7822} & \underline{0.6724} & 166.31 \\
\textcolor{pink!60!black}{\textbf{FreeUV}} & \textbf{0.8490} & \textbf{0.7559} & \textbf{142.39} & \textbf{0.8272} & \textbf{0.7948} & \textbf{153.43} & \textbf{0.7997} & \textbf{0.6835} & \underline{158.55} \\
\bottomrule
\end{tabular}
}}
\end{center}
\label{tab:quantitative_uv}
\end{table*}

To quantitatively evaluate our approach, we adopted the comparison methodology from UV-IDM~\cite{uvidm2024}, focusing on non-iterative texture refinement methods and excluding iterative techniques like FFHQ-UV. We employed metrics such as DINO-I~\cite{DINO2021}, CLIP-I~\cite{CLIP2021}, and FID~\cite{FID2017} to measure both the semantic alignment and visual quality between original 2D face images and the recovered UV textures. As shown in \cref{tab:quantitative_uv}, our method achieved superior results across DINO-I, CLIP-I, and FID scores, highlighting its capability to capture semantically meaningful and visually coherent features, while delivering textures that align closely with realistic distributions. These results underscore FreeUV’s robustness in preserving intricate facial details and ensuring high-quality, lifelike texture synthesis.

\begin{figure}[t]
    \centering
    \includegraphics[width=\linewidth]{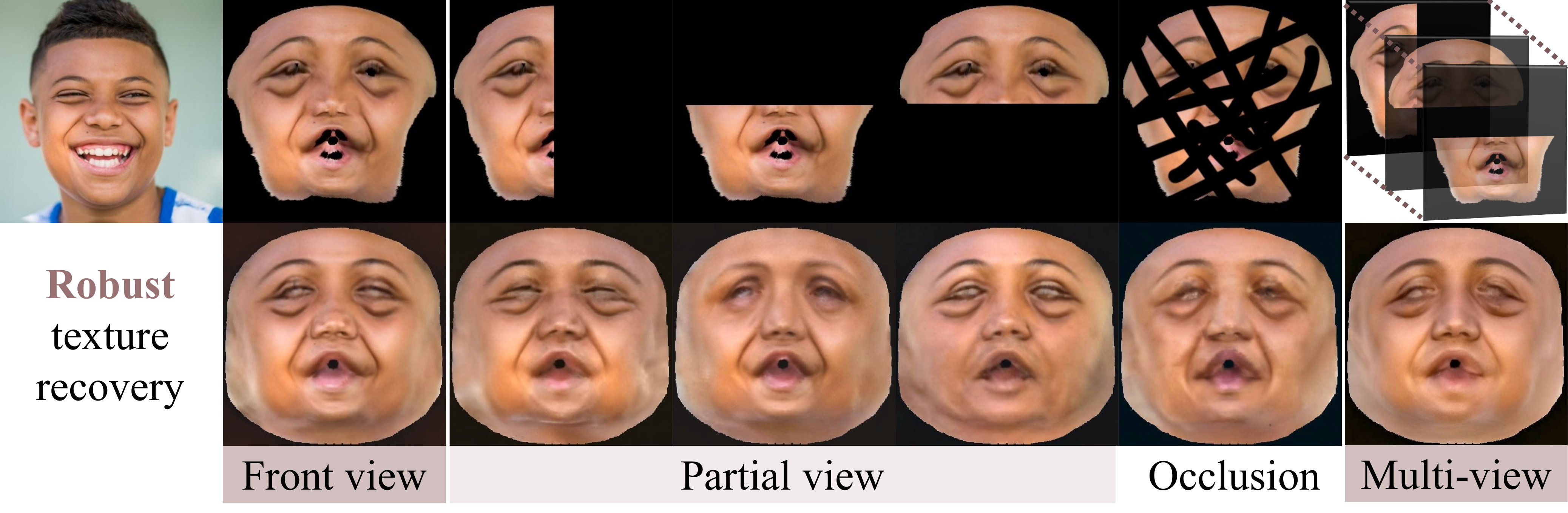}
    \caption{\textbf{Robustness evaluation.} Our method effectively handles partial views, seamlessly completing missing regions and maintaining color consistency, even with extensive occlusions.
}
\label{fig:result_recovery}
\end{figure}

\vskip0.5\baselineskip
\noindent \textbf{Robustness.}
\cref{fig:result_recovery} demonstrates the robustness of our method in handling partial and incomplete inputs. To simulate various input views, we manually masked sections of the unwrapped texture from a frontal face image. As shown in columns 3 to 5, each masked input serves as a partial view, where the unmasked portions retain detail and continuity, while the masked areas are seamlessly filled, maintaining color consistency and smooth transitions.

Even under extreme occlusion, as illustrated in column 6, our method generates visually coherent and credible results. Notably, when multiple partial views are concatenated as batch input to the Flaw-Tolerant Facial Detail Extractor, the model’s output closely approaches that of a complete frontal view. The network selectively integrates the best features from each partial view, resulting in a high-quality synthesis. This capacity to handle multiple partial views opens avenues for applying FreeUV to multi-view stereo and video-based texture recovery, directions we aim to explore in future work.

\begin{figure}[t]
    \centering
    \includegraphics[width=\linewidth]{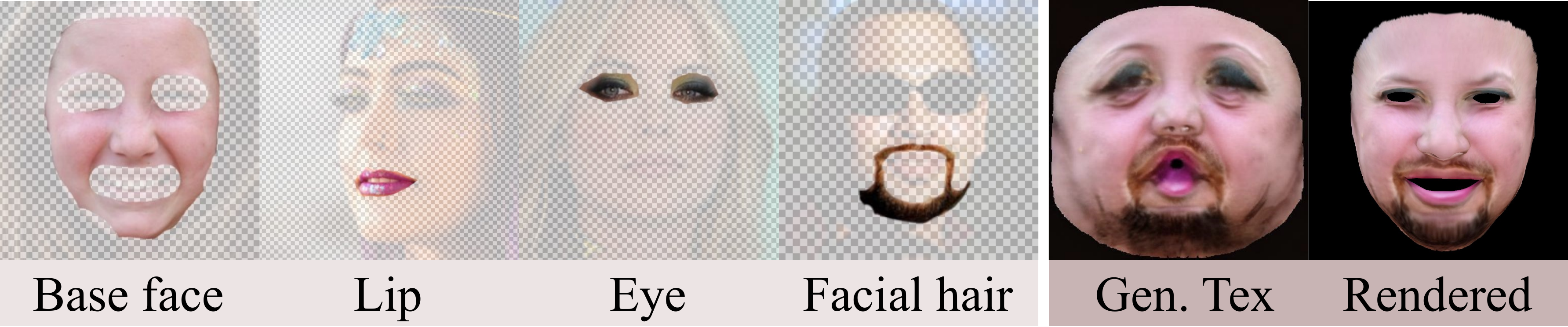}
    \caption{\textbf{Result of customized local editing application.} Our method enables seamless transfer of specific facial features from different source images onto a base face, creating a coherent UV texture that combines multiple attributes.
}
\label{fig:result_combine}
\end{figure}

\vskip0.5\baselineskip
\noindent \textbf{Customized local editing.}
Following insights gained from \cref{fig:result_recovery}, our method facilitates flexible, customized local editing. As demonstrated in \cref{fig:result_combine}, specific facial features—such as lips, eyes, and facial hair—can be selectively transferred from different source images to a base face. After unwrapping and layering these chosen regions into a single UV texture, the network completes the composition. This approach yields a coherent facial UV texture, seamlessly integrating the transferred features with the base face.
Our method excels at preserving natural color transitions and maintaining structural consistency, resulting in realistic edits that enhance the base image without visible seams or inconsistencies. This capability not only supports detailed, user-driven customization but also showcases FreeUV's adaptability to high-fidelity, personalized editing applications.

\begin{figure}[t]
    \centering
    \includegraphics[width=\linewidth]{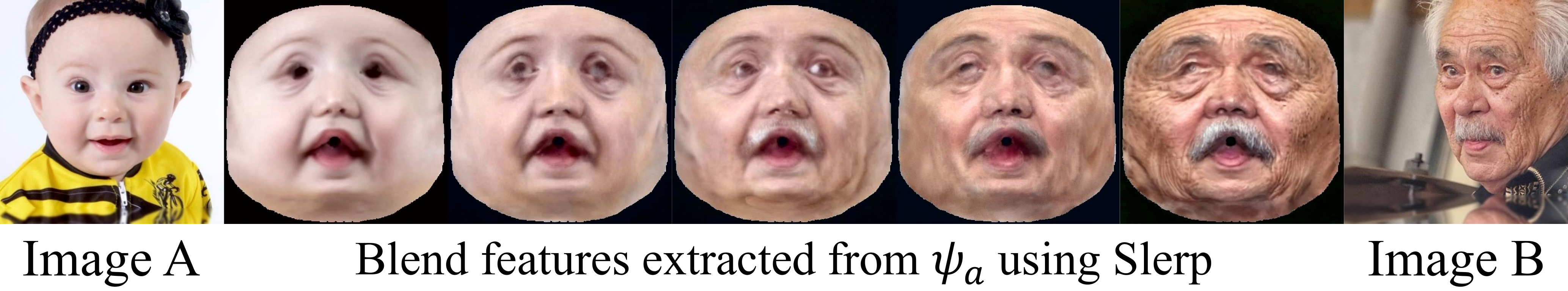}
    \caption{\textbf{Result of facial feature interpolation.} Facial feature blending is achieved by interpolating features from two input images, which can be used for, \eg, customizable transformations.
}
\label{fig:result_interpolate}
\end{figure}

\vskip0.5\baselineskip \noindent \textbf{Facial feature interpolation.}
By interpolating between two distinct input images, our method enables applications in facial feature blending. As shown in \cref{fig:result_interpolate}, we first extract features from Image A and Image B individually using our facial detail extractor. We then merge these features through a spherical linear interpolation (slerp) method, which provides smooth, visually coherent transitions between the two images. This approach supports applications such as 3D face aging, where facial features are blended to simulate realistic aging effects or other transformations. Notably, this interpolation captures subtle details and maintains structural integrity, allowing for highly customizable facial transformations with natural results.

\begin{figure}[t]
    \centering
    \includegraphics[width=\linewidth]{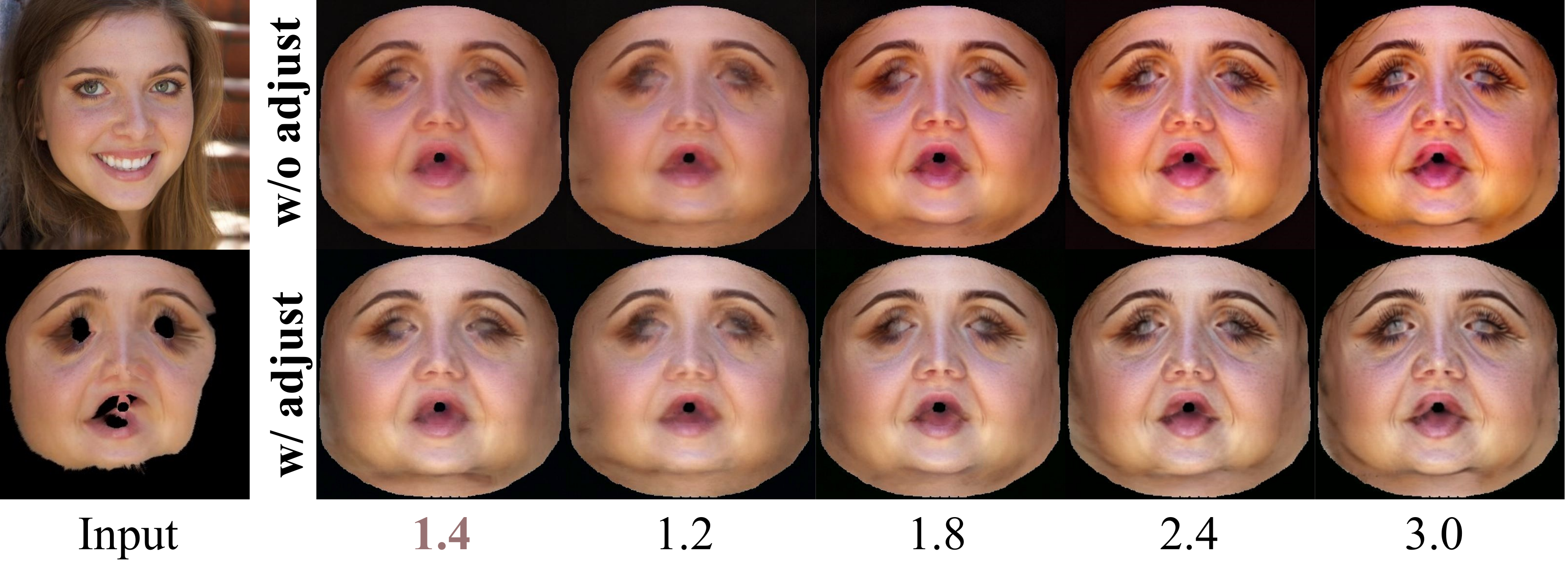}
    \caption{\textbf{
    Effect of classifier-free guidance (CFG) scale and subsequent color adjustment.
    } 
    The first row shows results with varying CFG scales, where lower scales reduce detail and higher scales amplify both detail and noise, potentially causing color mismatches. The second row applies Lab space color adjustment to the corresponding results from the first row, ensuring a coherent tone aligned with the original input.
    }
\label{fig:result_color_transfer}
\end{figure}

\subsection{Ablation Studies}
\vskip0.5\baselineskip
\noindent \textbf{Color adjustment.} 
\cref{fig:result_color_transfer} shows the results with varying classifier-free guidance scales and applying Lab space color adjustment. Lower guidance scales diminished the detail, while higher scales introduced excessive detail and noise, causing discrepancies with the original image. After testing, we found a guidance scale of 1.4 achieved optimal detail and natural appearance. With color adjustment, the final results exhibit a consistent tone and enhanced visual alignment with the original unwrapped input.

\begin{figure}[t]
    \centering
    \includegraphics[width=\linewidth]{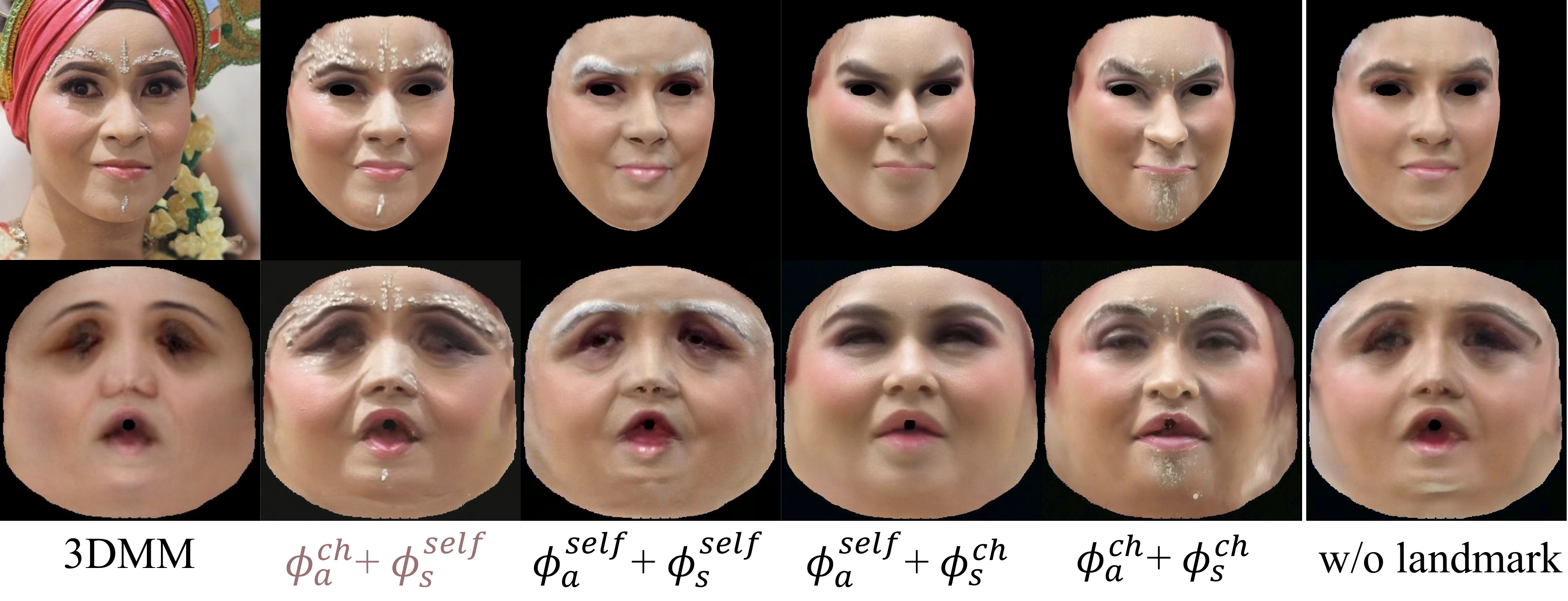}
    \caption{\textbf{Comparison of different module configurations in channel and self-attention for facial UV texture recovery.} Our approach shows that channel attention excels in capturing fine-grained details for UV-to-2D mapping, while self-attention better maintains structural alignment for 2D-to-UV mapping in the 3DMM domain.
}
\label{fig:result_ablation}
\end{figure}

\begin{table}
\caption{\textbf{Ablation study using different module configurations.} 
Here ``w/o lm.'' indicates ``without landmarks'' whereas ``w/o adj.'' means ``without color adjustment.''
}
\begin{center}
\resizebox{\linewidth}{!}{
{\small
\begin{tabular}{lcccccc}
\toprule
\multicolumn{1}{l}{\textbf{}} & \multicolumn{1}{r}{\textcolor{pink!60!black}{${\phi}_{a}^{ch}$+${\phi}_{s}^{self}$}} & \multicolumn{1}{r}{${\phi}_{a}^{self}$+${\phi}_{s}^{self}$} & \multicolumn{1}{r}{${\phi}_{a}^{self}$+${\phi}_{s}^{ch}$} & \multicolumn{1}{r}{${\phi}_{a}^{ch}$+${\phi}_{s}^{ch}$} & \multicolumn{1}{r}{w/o lm.} & \multicolumn{1}{r}{w/o adj.} \\
\midrule
\multicolumn{1}{r}{RMSE$\downarrow$} & \multicolumn{1}{r}{\textbf{0.0276}} & \multicolumn{1}{r}{0.0302} & \multicolumn{1}{r}{0.0367} & \multicolumn{1}{r}{0.0379} & \multicolumn{1}{r}{0.0292} & \multicolumn{1}{r}{\underline{0.0282}} \\
\multicolumn{1}{r}{SSIM$\uparrow$} & \multicolumn{1}{r}{\textbf{0.8001}} & \multicolumn{1}{r}{0.7881} & \multicolumn{1}{r}{0.7876} & \multicolumn{1}{r}{0.7648} & \multicolumn{1}{r}{0.7928} & \multicolumn{1}{r}{\underline{0.7992}} \\
\multicolumn{1}{r}{LPIPS$\downarrow$} & \multicolumn{1}{r}{\textbf{0.0463}} & \multicolumn{1}{r}{\underline{0.0474}} & \multicolumn{1}{r}{0.0539} & \multicolumn{1}{r}{0.0639} & \multicolumn{1}{r}{0.0481} & \multicolumn{1}{r}{0.0531} \\
\multicolumn{1}{r}{PSNR$\uparrow$} & \multicolumn{1}{r}{\textbf{30.848}} & \multicolumn{1}{r}{30.397} & \multicolumn{1}{r}{28.693} & \multicolumn{1}{r}{28.417} & \multicolumn{1}{r}{30.624} & \multicolumn{1}{r}{\underline{30.828}} \\
\bottomrule
\end{tabular}
}}
\end{center}
\label{tab:ablation_render}
\end{table}

\vskip0.5\baselineskip
\noindent \textbf{Module selection.}
We assessed the roles of channel attention and self-attention in our task by combining different facial detail extractors in networks ${\phi}_{a}$ and ${\phi}_{s}$. As shown in \cref{fig:result_ablation}, our experiments reveal that, for the UV-to-2D mapping in the in-the-wild domain, using channel attention in appearance network ${\phi}_{a}$ (${ \phi}_{a}^\mathit{ch}$) effectively preserves realistic details. This configuration captures fine-grained features, whereas self-attention (${\phi}_{a}^\mathit{self}$) tends to smooth the details, yielding a slightly blurred outcome. In one case, channel attention successfully retained delicate facial adornments, while self-attention produced a less detailed, more uniform result.

On the other hand, for the 2D-to-UV mapping in the 3DMM domain, self-attention in structure network ${\phi}_{s}$ (${\phi}_{s}^\mathit{self}$) better maintains structural alignment with the 3DMM UV layout, while channel attention (${\phi}_{s}^\mathit{ch}$) introduces distortions that disrupt this consistency. Consequently, rendered results using channel attention in this setting may not align accurately with the original image.

Furthermore, excluding 2D landmarks as alignment aids in appearance network ${\phi}_{a}$ results in notable detail loss, likely due to structural inaccuracies from imperfect 3DMM fitting. These inaccuracies hinder precise alignment of structure and appearance. Including detected 2D landmarks from the image mitigates this issue, improving detail preservation in the generated UV textures.

As shown in \cref{tab:ablation_render}, we conducted quantitative experiments on different module combinations by rendering the recovered UV textures for comparison with the original 2D face image. Key evaluation metrics included RMSE, SSIM, LPIPS~\cite{lpips}, and PSNR. Our selected configuration achieved the top scores, with the results prior to color adjustment ranking as the second best. These findings indirectly confirm the effectiveness of our selected attention modules.

\subsection{Limitation}
As illustrated in \cref{fig:result_ablation}, while our method successfully recovers highly detailed features, it faces limitations with very fine-grained elements, such as facial accessories, spots, and blemishes. These details may exhibit slight shifts in position or quantity. Additionally, the network struggles to precisely localize these features within specific areas; for instance, when reconstructing regions originally covered by a hat, the network may unintentionally extend surrounding details to ensure texture continuity, which can compromise the accuracy of localized texture recovery.

\section{Conclusions}
We introduced FreeUV, a novel framework for 3D facial UV texture recovery that operates without the need for ground-truth UV data. By leveraging a pre-trained diffusion model and a Cross-Assembly inference strategy, FreeUV disentangles appearance from structure, enabling precise capture of fine-grained textures alongside robust structural integrity. This approach allows FreeUV to generate realistic, high-quality UV textures directly from single-view, in-the-wild images, handling complex details with adaptability across varied real-world conditions. Furthermore, FreeUV’s versatility extends to applications including customized local editing, facial feature interpolation, and multi-view texture recovery. By reducing data dependency while maintaining texture fidelity, FreeUV offers a scalable solution for high-quality facial texture recovery.

%% file: sec/X_suppl.tex
\clearpage
\setcounter{page}{1}
\maketitlesupplementary

\renewcommand{\thefigure}{S.\arabic{figure}}
\renewcommand{\thetable}{S.\arabic{table}}
\renewcommand{\theequation}{S.\arabic{equation}}

\appendix

In this supplementary document, we first describe the training data preparation process, including the generation of masked data and its use in both the in-the-wild and 3DMM domains (\cref{sec:training_data_preparation}). Then, we present ablation studies on network structures to analyze their impact on the 3DMM UV structure and realism (\cref{sec:ablation_studies}). Finally, we provide additional qualitative comparisons with existing methods, showcasing the superior performance of our approach (\cref{sec:additional_results}).

\begin{figure*}[t]
    \centering
    \includegraphics[width=0.96\linewidth]{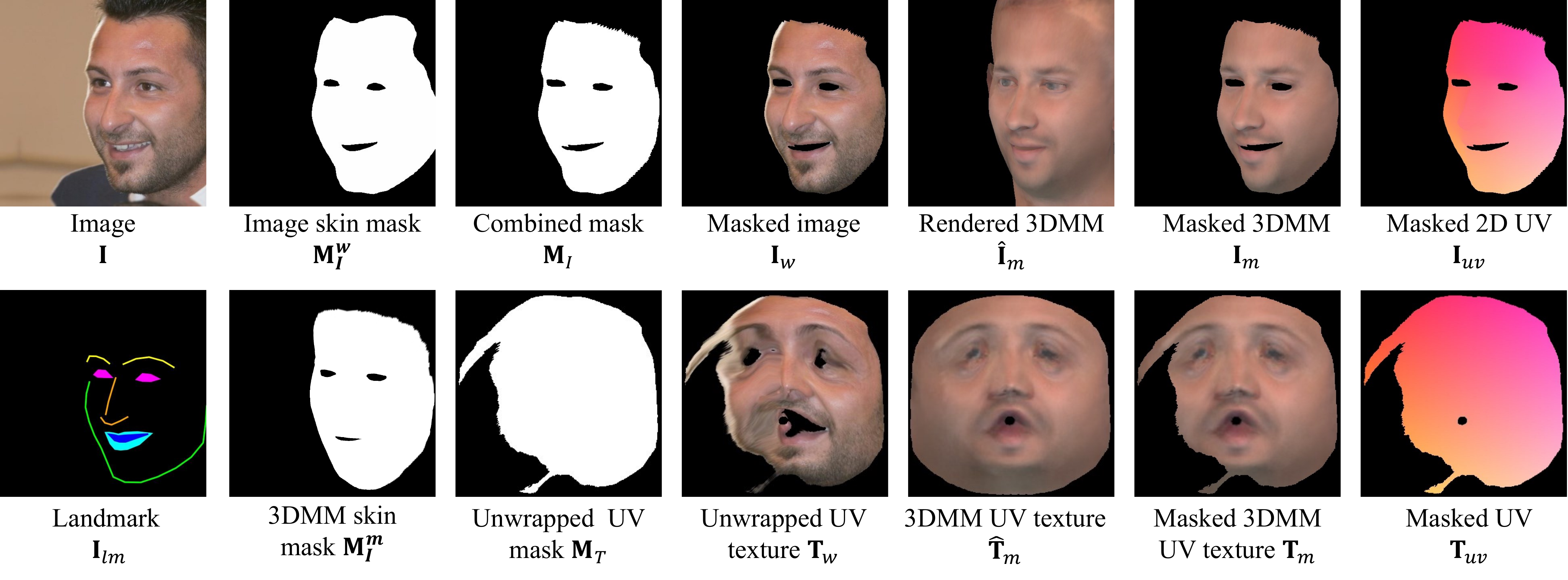}
    \caption{\textbf{Training Data Preparation Process.} From a single image \( \mathbf{I} \), this process produces a dataset suitable for training.
    }
\label{fig:suppl_data_prepare}
\end{figure*}

\section{Training Data Preparation}
\label{sec:training_data_preparation}

Our training data preparation process is illustrated in \cref{fig:suppl_data_prepare}. Given a face image \( \mathbf{I} \), we first derive the skin region mask \( \mathbf{M}_{I}^{w} \) using a face segmentation algorithm~\cite{BiSeNet2018}. Additionally, we extract another skin region mask \( \mathbf{M}_{I}^{m} \) from the reconstructed 3DMM skin area.
These two masks are combined via element-wise multiplication to produce the final mask \( \mathbf{M}_I \), which represents the overlapping skin regions shared by both the in-the-wild image and the 3DMM-generated image. Applying this combined mask to the original image \( \mathbf{I} \) yields the masked in-the-wild image \( \mathbf{I}_w \). For both in-the-wild and 3DMM data, we utilize masked data to ensure consistency during the inference stage integration.

For 3D face reconstruction, we employ the FLAME 3DMM model~\cite{FLAME}. We directly utilize the reconstruction method proposed in~\cite{MakeupExtract, makeupPriors2024} without additional training. The method in~\cite{MakeupExtract, makeupPriors2024} is a re-trained version of the Deep3Dface method~\cite{deng2020accurate}, specifically adapted for the FLAME model. The reconstructed 3DMM result is denoted as \( \hat{\mathbf{I}}_m \). Subsequently, we isolate the skin region from the UV texture by cropping and enlarging it, producing \( \hat{\mathbf{T}}_m \).
Using the reconstructed 3DMM shape, we sample pixels from the masked image \( \mathbf{I}_w \) and unwrap them into UV space, resulting in the unwrapped UV texture \( \mathbf{T}_w \). Similarly, the combined mask \( \mathbf{M}_I \) is unwrapped to generate the UV mask \( \mathbf{M}_T \). By applying the UV mask to the UV texture and UV position map from the 3DMM, we obtain the masked UV texture \( \mathbf{T}_m \) and the UV position map \( \mathbf{T}_{uv} \), respectively. These masked outputs are then reprojected into 2D space, resulting in the 2D images \( \mathbf{I}_m \) and \( \mathbf{I}_{uv} \).

This pipeline, starting from a single input image \( \mathbf{I} \), produces a dataset suitable for training. However, it is notable that generating a complete and realistic UV texture is challenging. The accuracy of the unwrapped UV texture \( \mathbf{T}_w \) is highly dependent on the performance of 3D face reconstruction. Flaws such as inaccuracies in 3DMM fitting and self-occlusion often result in distorted textures with missing regions.

\begin{figure}[t]
    \centering
    \includegraphics[width=\linewidth]{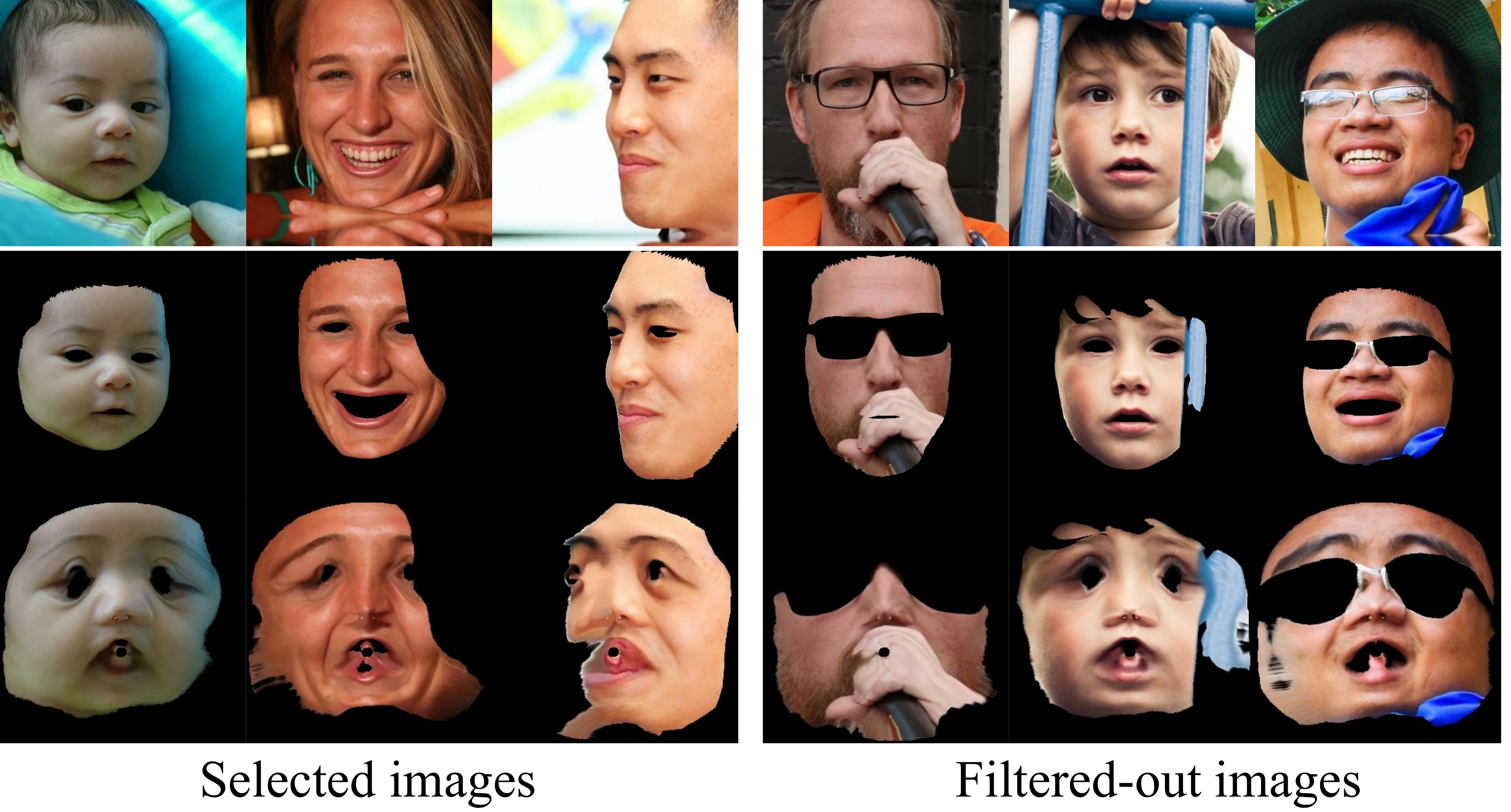}
    \caption{\textbf{Examples of Final In-the-Wild Images Selected for Training.} Images with segmentation failures—such as those containing residual artifacts from hands or other objects—were excluded to ensure data quality.}

\label{fig:suppl_data_select}
\end{figure}

\begin{figure}[t]
    \centering
    \includegraphics[width=\linewidth]{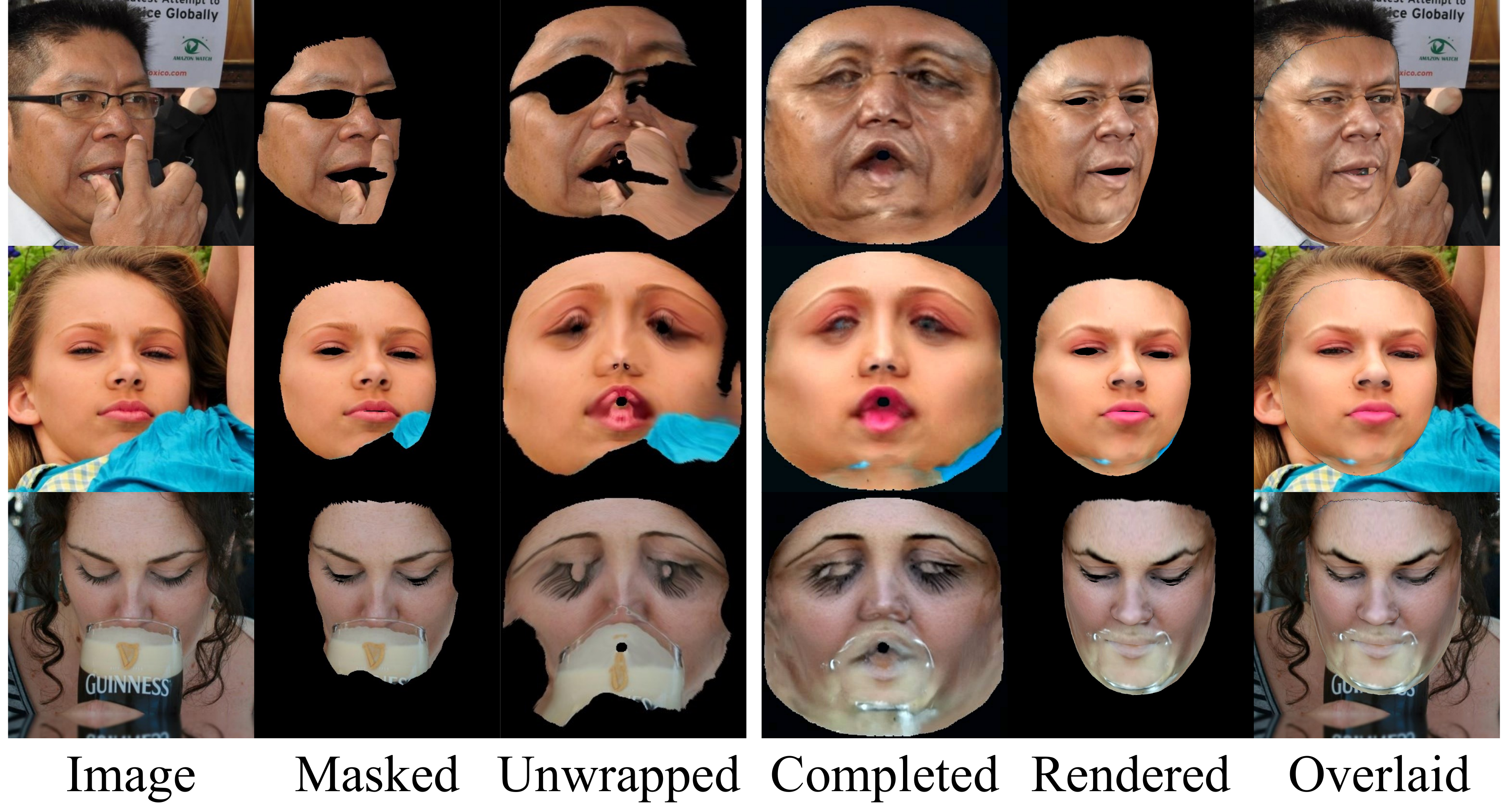}
    \caption{\textbf{Impact of Segmentation Failures on Inference Results.} Failed segmentation in input images can degrade output quality. Advanced face segmentation techniques will be helpful in addressing this issue.}

\label{fig:suppl_object_input_result}
\end{figure}

Examples of the final in-the-wild images selected for training are shown in \cref{fig:suppl_data_select}. Images with failed face segmentation—such as those containing residual artifacts from hands or other objects—were manually excluded, as they do not accurately correspond to the skin region of the human face. During the inference stage, images with failed segmentation used as input can degrade the quality of the final output, as shown in \cref{fig:suppl_object_input_result}. To address this issue, we plan to leverage more advanced face segmentation techniques in the future to ensure robust performance in such cases. Nevertheless, our method is still capable of partially alleviating the impact of occlusions.

\section{Additional Networks on Ablation Studies}
\label{sec:ablation_studies}

\begin{figure}[t]
    \centering
    \includegraphics[width=\linewidth]{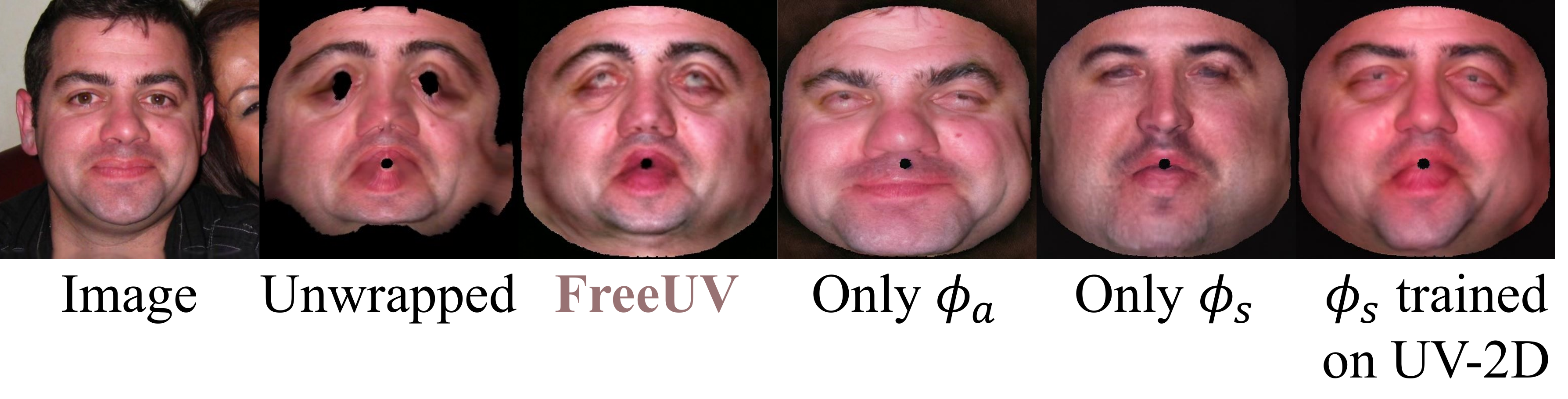}
    \caption{\textbf{Ablation Studies on Network Structures.} Results using only the appearance network \( \phi_a \), only the shape network \( \phi_s \), and FreeUV with different configurations highlight the challenges in preserving the structure of the 3DMM UV map and maintaining realism.}

\label{fig:suppl_ablation}
\end{figure}

Since both appearance network ${\phi}_{a}$ and structure network ${\phi}_{s}$ can independently generate UV texture, we evaluated their outputs to assess their ability to preserve the structure and realism of the 3DMM UV map. The results are shown in the 4th and 5th columns of \cref{fig:suppl_ablation}. Their inputs are the same as those of inference network ${\phi}_{i}$, \ie, the unwrapped UV texture \( \mathbf{T}_w \) and the complete UV position map $\mathbf{\Upsilon}_w$. Neither of ${\phi}_{a}$ and ${\phi}_{s}$ is capable of preserving the structure of the 3DMM UV map. Structure network ${\phi}_{s}$, trained on the 3DMM data domain, , fails to maintain consistency with the input UV map and introduces distortions.

We also observed that structural degradation still occurs when the structure network \( \phi_s \) is trained using a UV-to-2D input-output configuration (see the 6th column of \cref{fig:suppl_ablation}), even though this setup aligns with the consistent pattern of the 3DMM data domain.

In contrast, FreeUV adopts a UV-to-UV Cross-Assembly Inference strategy, ensuring consistent alignment and structural integrity within the UV space. This strategy effectively allows the appearance network \( \phi_a \) and the structure network \( \phi_s \) to complement each other's strengths.

\section{Comparison with DSD-GAN}
\begin{figure}[h]
    \centering
    \includegraphics[width=0.96\linewidth]{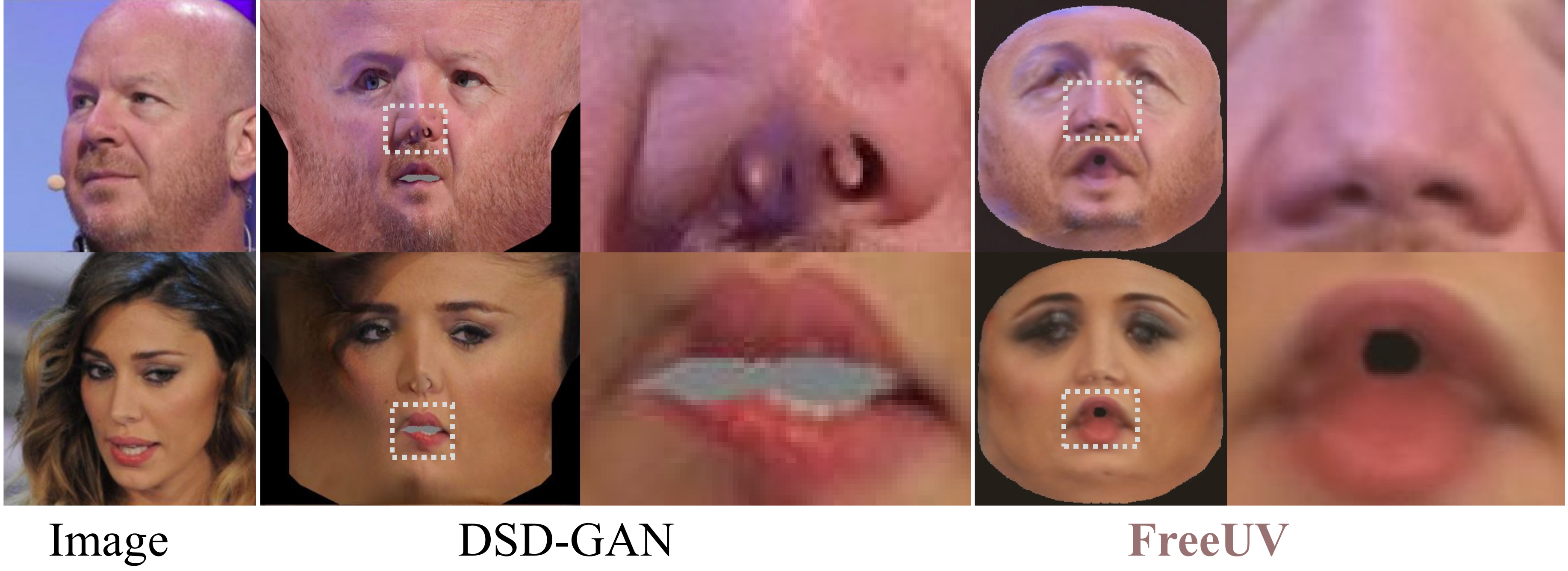}
    \caption{\textbf{Comparison with DSD-GAN.} Our method shows improved structural alignment and texture fidelity.}
\label{fig:result_dsd}
\end{figure}

To the best of our knowledge, DSD-GAN~\cite{dsd-gan} is the first and only method to achieve complete UV texture completion without relying on ground truth UV data. It employs a dual-space discriminator GAN, applying two discriminators in both UV map space and image space to learn facial structures and texture details. Under the same UV ground-truth-free setting, our key contribution lies in improving structural consistency and texture fidelity. Furthermore, our method leverages a pre-trained diffusion model, enhancing robustness to out-of-domain scenarios. As shown in \cref{fig:result_dsd}, DSD-GAN exhibits misalignment artifacts, particularly in the nasal and lip regions, as observed in their paper (since the official code is not publicly available).

\section{Additional Results}
\label{sec:additional_results}
\begin{figure*}[t]
    \centering
    \includegraphics[width=\linewidth]{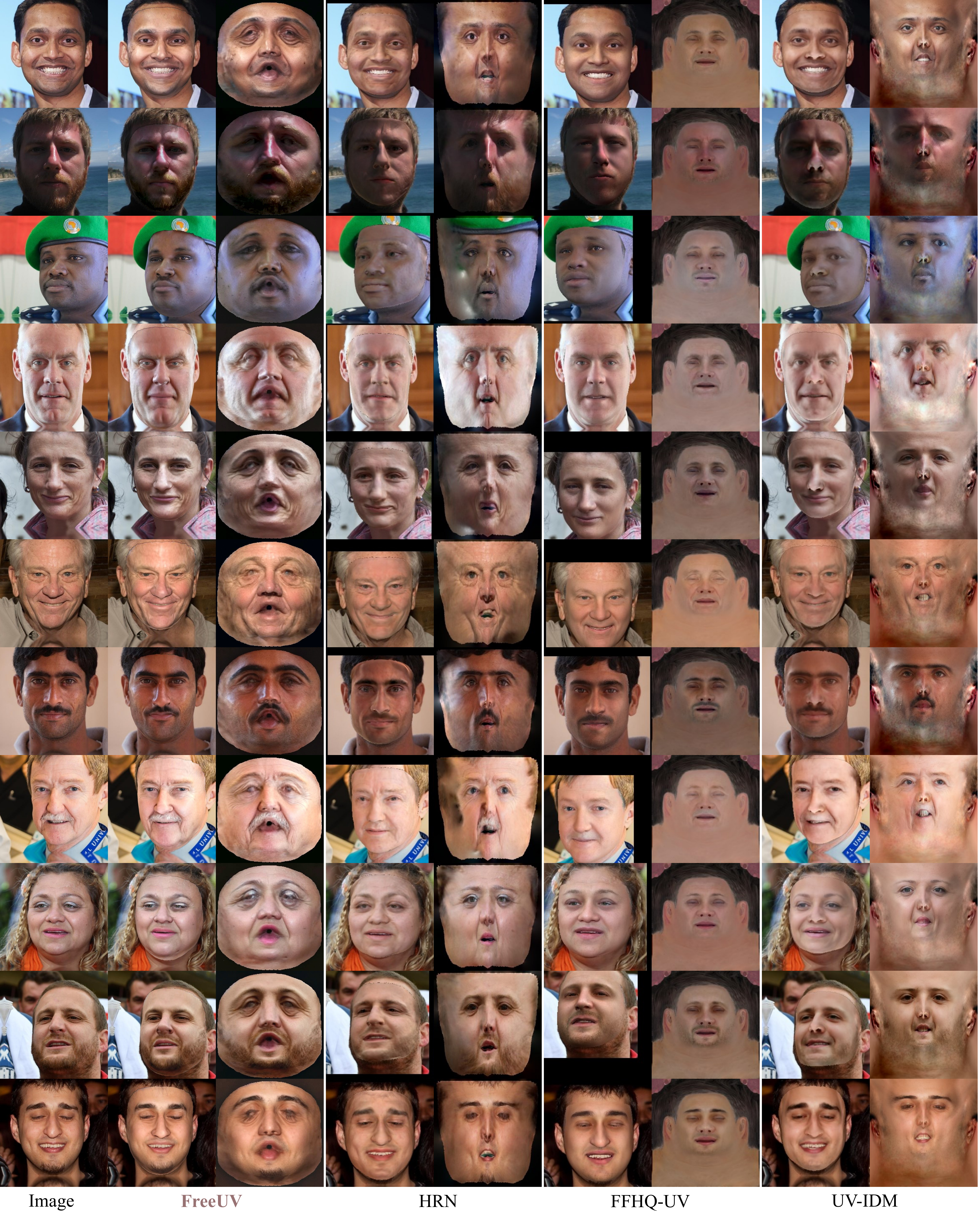}
    \caption{\textbf{Comparison of Results.} Our method outperforms HRN~\cite{hrn2023}, FFHQ-UV~\cite{FFHQ-UV}, and UV-IDM~\cite{uvidm2024} in capturing fine details, achieving realism, and maintaining robustness.}
\label{fig:suppl_result_1}
\end{figure*}

\begin{figure*}[t]
    \centering
    \includegraphics[width=\linewidth]{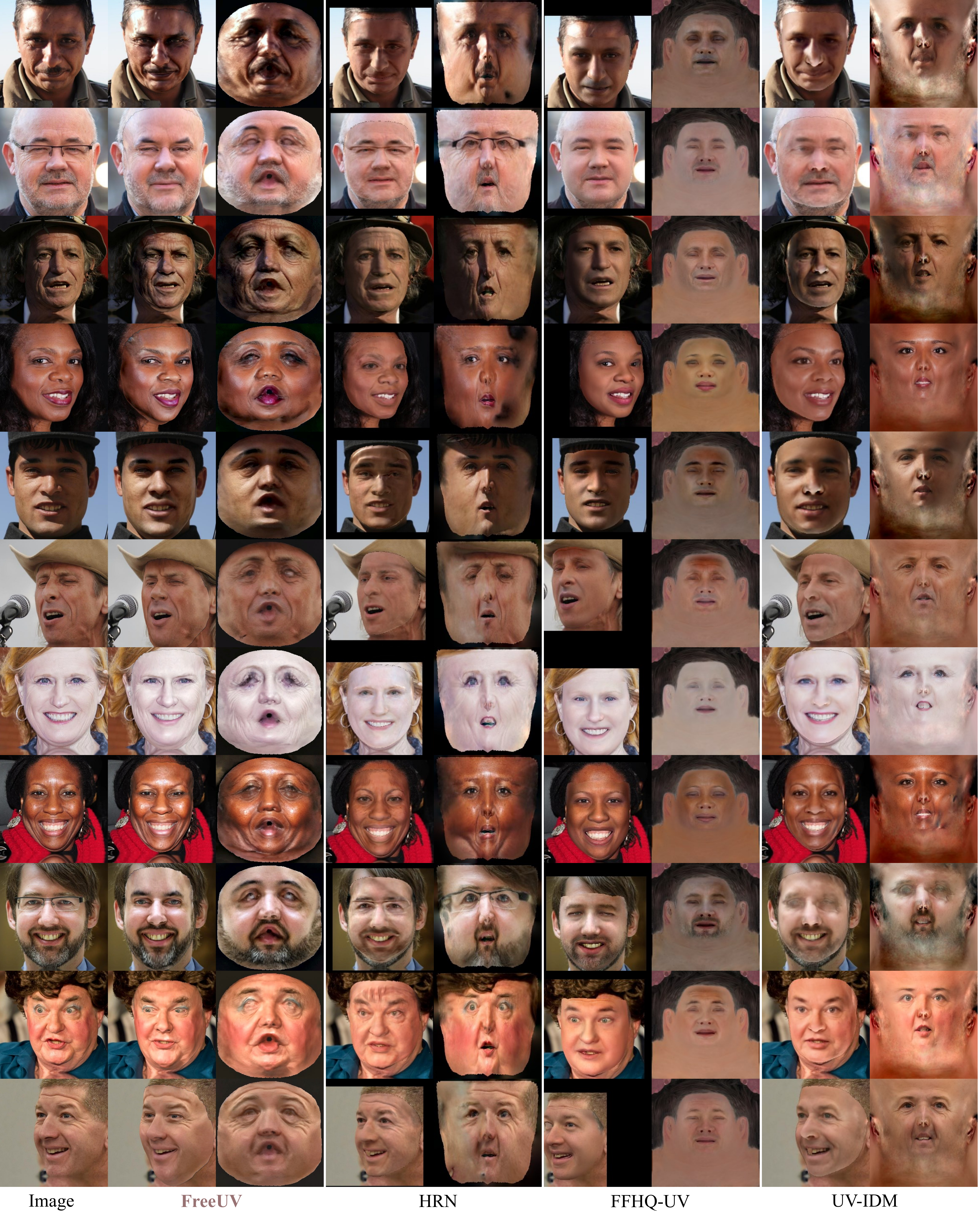}
    \caption{\textbf{Comparison of Results.} Our method outperforms HRN~\cite{hrn2023}, FFHQ-UV~\cite{FFHQ-UV}, and UV-IDM~\cite{uvidm2024} in capturing fine details, achieving realism, and maintaining robustness.}
\label{fig:suppl_result_2}
\end{figure*}

\begin{figure*}[t]
    \centering
    \includegraphics[width=\linewidth]{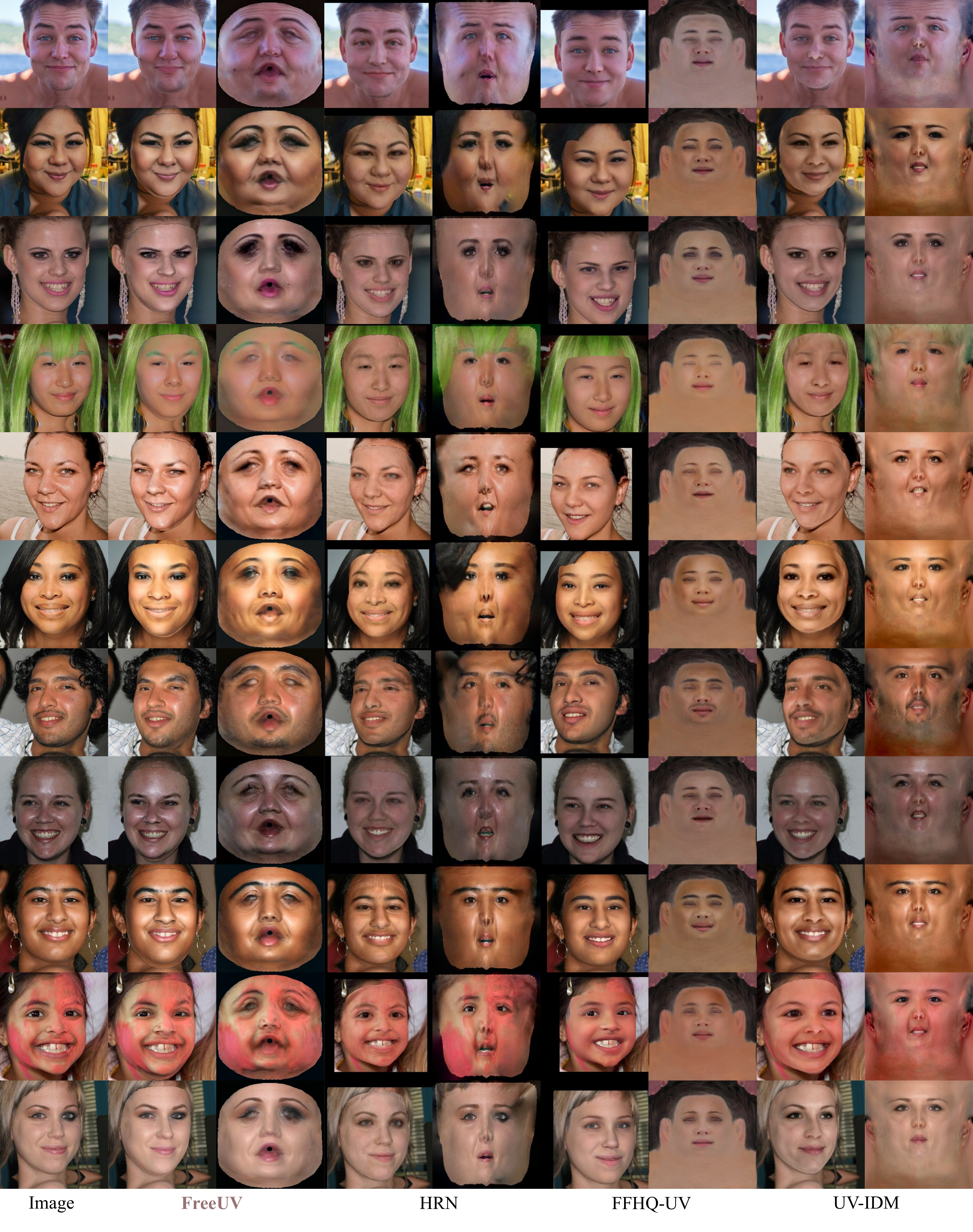}
    \caption{\textbf{Comparison of Results.} Our method outperforms HRN~\cite{hrn2023}, FFHQ-UV~\cite{FFHQ-UV}, and UV-IDM~\cite{uvidm2024} in capturing fine details, achieving realism, and maintaining robustness.}
\label{fig:suppl_result_3}
\end{figure*}

We present additional results in Figs.~\ref{fig:suppl_result_1}, \ref{fig:suppl_result_2}, and \ref{fig:suppl_result_3}. Compared to HRN~\cite{hrn2023}, FFHQ-UV~\cite{FFHQ-UV}, and UV-IDM~\cite{uvidm2024}, our method excels in capturing fine details, achieving realistic outputs, and demonstrating enhanced robustness, particularly in preserving features such as beards, wrinkles, specular highlights, and makeup.

